%% file: main.tex
\documentclass[conference]{IEEEtran}
\usepackage{cite}
\usepackage{amsmath,amssymb,amsfonts}
\usepackage{algorithmic}
\usepackage{graphicx}
\usepackage{subfigure}
\usepackage{textcomp}
\usepackage{xcolor}
\usepackage{url}
\usepackage[whole]{bxcjkjatype}
\def\BibTeX{{\rm B\kern-.05em{\sc i\kern-.025em b}\kern-.08em
    T\kern-.1667em\lower.7ex\hbox{E}\kern-.125emX}}
\usepackage[binary-units=true]{siunitx}
\newcommand{\figref}[1]{\figurename~\ref{#1}}
\newcommand{\tabref}[1]{\tablename~\ref{#1}}
\begin{document}
\setlength{\parskip}{0.5\baselineskip}
\title{Building a Computer Mahjong Player via Deep Convolutional Neural Networks
}

\author{\IEEEauthorblockN{Shiqi Gao}
\IEEEauthorblockA{\textit{Graduate School of Information Science and Techonology}\\
\textit{The University of Tokyo}\\
7-3-1 Hongo, Bunkyo-ku, Tokyo, Japan\\
gaosq0604@akg.t.u-tokyo.ac.jp}
\and
\IEEEauthorblockN{Fuminori Okuya}
\IEEEauthorblockA{\textit{Graduate School of Information Science and Techonology}\\
\textit{The University of Tokyo}\\
7-3-1 Hongo, Bunkyo-ku, Tokyo, Japan\\
okuya23@akg.t.u-tokyo.ac.jp}
\and
\IEEEauthorblockN{Yoshihiro Kawahara}
\IEEEauthorblockA{\textit{Graduate School of Information Science and Techonology}\\
\textit{The University of Tokyo}\\
7-3-1 Hongo, Bunkyo-ku, Tokyo, Japan\\
kawahara@akg.t.u-tokyo.ac.jp}
\and
\IEEEauthorblockN{Yoshimasa Tsuruoka}
\IEEEauthorblockA{\textit{Graduate School of Engineering}\\
\textit{The University of Tokyo}\\
7-3-1 Hongo, Bunkyo-ku, Tokyo, Japan\\
tsuruoka@logos.t.u-tokyo.ac.jp}
}

\maketitle

\begin{abstract}
The evaluation function for imperfect information games is always hard to define but owns a significant impact on the playing strength of a program. Deep learning has made great achievements these years, and already exceeded the top human players' level even in the game of Go. In this paper, we introduce a new data model to represent the available imperfect information on the game table, and construct a well-designed convolutional neural network for game record training. We choose the accuracy of tile discarding which is also called as the agreement rate as the benchmark for this study. Our accuracy on test data reaches 70.44\%, while the state-of-art baseline is 62.1\% reported by Mizukami and Tsuruoka (2015), and is significantly higher than previous trials using deep learning, which shows the promising potential of our new model. For the AI program building, besides the tile discarding strategy, we adopt similar predicting strategies for other actions such as stealing (pon, chi, and kan) and riichi. With the simple combination of these several predicting networks and without any knowledge about the concrete rules of the game, a strength evaluation is made for the resulting program on the largest Japanese Mahjong site `Tenhou'. The program has achieved a rating of around 1850, which is significantly higher than that of an average human player and of programs among past studies.
\end{abstract}


\section{Introduction}
The studies on Artificial Intelligence (AI) players on perfect information games are highly advanced and their strength has already exceeded the level that no human beings can touch even in the game of Go, one of the most complex perfect information games, by the AI AlphaGo series in the year of 2017.

Compared with perfect information games which refer that each player has the same information, there is still a long way to go for imperfect information ones since they are more complex and the players need to balance all possible outcomes while making a decision. It is still very difficult for AI to find the best strategy when faced with situations of imperfect information.

The usual solution for imperfect information game solving is to achieve a Nash equilibrium situation~\cite{nash}. The Nash equilibrium is a proposed solution for multi-player non-cooperative games in which each player is rational enough and no one will gain any additional benefits by only changing his own strategy. There has been a breakthrough of approaching Nash equilibrium points in no-limit Texas Hold'em~\cite{libratus, libratus2}, a two-player unlimited form of poker by using subgame solving and Counterfactual Regret Minimization (CFR)~\cite{cfr}. However, for the game of Mahjong, it has more players and much larger exploring space and complexity, so we aim at other solutions.

In this paper, we choose mahjong as our research tool due to its large complexity and popularity. Japanese Mahjong has a strict frame of rules for competitions, together with many high-level past game records which are called `haifu' for the network to study. Agreement rate, the rate of the network choosing the same strategy with the real data records, is a very important benchmark for estimating the situation of the game which is also called the evaluation function. There are many traditional AI methods for the game of Mahjong, but they are mainly built by artificially extracting features and designing function blocks. By contrast, deep learning is recognized as the next generation method for game AI for its ability of automatic feature extraction and learning. Although there have been related studies in full-connected neural networks~\cite{deeplearningmahjong} and CNN~\cite{cnnmahjong}, they still cannot exceed the performance of traditional methods' due to their own limitations in both data structure and network structure.

We propose a new method by using deep convolutional neural networks for this task. We design a new data structure with three dimensions in order to contain the information available on the table, and make fine training. We focus on the benchmark of agreement rate, and try to achieve a best accuracy for game record learning. We also build an AI program which directly combines several trained neural networks together for strength evaluation, and show great potential of our model.

The paper is organized as follows. Section~\uppercase\expandafter{\romannumeral2} gives basic introductions about rules and terms for the game of Japanese Mahjong. Section~\uppercase\expandafter{\romannumeral3} introduces related studies and works. Section~\uppercase\expandafter{\romannumeral4} provides a clear frame about our data model design and section~\uppercase\expandafter{\romannumeral5} mentions corresponding network structure design for discard and other action strategies. Section~\uppercase\expandafter{\romannumeral6} and~\uppercase\expandafter{\romannumeral7} shows training experiment and results. Section~\uppercase\expandafter{\romannumeral8} shares several interesting findings and finally section~\uppercase\expandafter{\romannumeral9} makes a conclusion.

\section{Basic Rules and Terms of Japanese Mahjong}
Mahjong is a tile-based imperfect information game which is commonly played by four players. Japanese Mahjong, also known as Riichi Mahjong, is a popular variation. Japanese Mahjong has 136 tiles with 34 types and four same tiles in each type. The 34 types are three kinds of number tiles of nine numbers each: 1m~(man)--9m, 1p~(pin)--9p, 1s~(sou)--9s, and seven honor tiles: East, South, West, North, Haku, Hatsu and Chun. The tiles are mixed and then arranged into four walls that are each two stacks high and 17 tiles wide. 26 of the stacks are used for starting hands, while 7 are used for a dead wall and the remaining 35 stacks form the playing wall~\cite{mahjongwiki}. Each player will have 13 tiles in hand, and take turns drawing a tile from the wall to form a 14-tile pattern. After that, he needs to choose one tile in hand to discard. The basic winning tile combination is $x\mathrm{(AAA)}+y\mathrm{(ABC)}+\mathrm{DD}$ while $x+y=4$, where AAA represents for a triplet (three same tiles, which is also called as set/mentsu), ABC represents for a sequence (three number tiles with continuing numbers, which is also called as run/shuntsu) and DD represents for a pair (two same tiles). The player can also make stealing when others discard specific tiles. The player can call a pon if he already has two same tiles in hand as the discarded one or he can call a chi if he can form a sequence with this tile.

Normally, four players initialize a game with a score of 25000 each, and one game of Mahjong usually contains four or eight parts of subgames. The rank is decided by players' scores at last.

Japanese Mahjong has several new features compared with original mahjong games.

\subsection{Riichi}
Riichi declaration is a ready hand announcement and is also a kind of yaku. A player can declare riichi to inform all other players that he is preparing to win and just needs one last tile to form a legal hand pattern. One can only declare riichi when he does not have any stealing tiles called previously. Once declaring riichi, one's tiles in hand cannot be changed anymore unless he has a closed kan or plus kan opportunity which is rare though, but he will own big chance of gaining larger scores once winning, since uradoras will count at that time.
\subsection{Yaku}
Yakus are specific tile patterns or conditions that possess values. For the game of Japanese Mahjong, at least one yaku is needed to be formed in hand in order for winning. Each yaku has a specific hand value called han which devotes to the hand tile score in an exponential way. Doras in hand cannot be regarded as yakus for the winning judgement, but will also devote to hand values once winning.
\subsection{Dora}
The hand tile pattern and the number of dora tiles in hand are two important factors for a big winning. Dora is a kind of bonus tile that enhances han value while winning. Each time a new subgame begins, there will be one dora tile known, since the dora indicator tile located atop the third stack from the last will be flipped over and revealed face up to all players. During the game, another dora indicator will be shown each time a kan is called, starting from the fourth stack from the last. It can have at most four Kans called among one subgame. Normally, the next tile of the dora indicator is regarded as the dora type. However, this is not always a linear relationship but divided into five smaller loops of 1m-9m, 1p-9p, 1s-9s, East-North and Haku-Chun. Among each loop, the succeeding tile of dora indicator becomes a dora, for example the next tile of 9m is not 1p but 1m, the next tile of North is not Haku but East. Aka five is another type of specific dora. Compared with dora indicators, aka five tiles always exist on table, and each one devotes to one han value. The game will have three red five tiles which are regarded as aka five doras, one each for man type, pin type, and sou type respectively. Different from normal five number tiles, they are in red color, while having all the functions of a normal tile.
\subsection{Uradora}
There is an uradora indicator under each dora indicator among the same tile stack. Uradoras will be revealed and counted only when one player achieves a winning after his declaring riichi beforehand, and what the uradora indicators is cannot be known until the final revealing, so it can be thought as an uncertain potential bonus only for riichi-declared players.
\subsection{Wind}
During the game, among all 34 kinds of tiles, we have seven types of honor tiles: East, South, West, North, Haku, Hatsu and Chun. For each subgame, there are two types of wind tiles that are very important to all players: the round wind and the own wind. The round wind is set same to all players in each subgame, initialized as East and will change as the game continues. The own wind, however, is different from player to player in one subgame. Not like Haku, Hatsu and Chun that any triplet of them is already regarded as one yaku along with one han value, among four types of wind tiles only the round wind and the player's own wind devote to yaku calculation.
\subsection{Winning}
There are mainly two types of winning, either to win from the wall or to win from a discard. Winning from the wall is also called self-drawn or tsumo, when the player gets his own tile picked up which forms a legal hand. While facing a self-drawn or tsumo, the other three players need to pay the winner some amount of score together. When one player catches a tile discarded by others and can form a legal hand with that, it will lead to a ron, and the original owner who discards that tile needs to pay the losing score entirely by himself. The dealer, whose own wind is East, always pays more and earns more. Once the dealer wins, he will earn the score 50\% larger, no matter by self-drawn or others' discard tiles. On the other hand, if another player gets a winning with self-drawn, the dealer will need to pay the score twice as many as the other two players.

\section{Related Works}
A one-on-one version of Texas Hold'em, one of the most popular variations of poker, has been solved by using subgame solving to reach a near Nash equilibrium point~\cite{libratus, libratus2}. However, there are still no suitable approaches to multi-player poker solutions for the unsustainable computational cost. Game of Mahjong, theoretically, has even larger complexity than the game of Texas Hold'em. With existing techniques and skills, it is impossible to reach the Nash equilibrium point for this game.

Traditional mahjong AIs are usually elaborately designed into several function blocks, such as the offense part and the defense part. For the offense part, the AI focuses on tile efficiency in order to make faster winning and larger scores, no matter what hand tiles other players have. It is acted as a one-player mahjong game. For the defense part, the goal is to make oneself always safe and discard safe tiles to avoid ron (which means defeated) by others. The training procedure is trying to achieve a suitable bias between them and decide strategies in different situations~\cite{mizukamijp, mizukamien}. The state-of-art agreement rate accuracy for game record learning on test dataset is 62.1\%~\cite{mizukamijp}.

Deep learning becomes hot these years and is featured for its strong automatic feature extraction ability without needing any artificial extraction, together with surprising learning capability. However, how to design a suitable data structure and build the corresponding neural network is still a problem. Tsukiji and Shibaraha~\cite{deeplearningmahjong} had a trial using a full-connected neural network in two layers and an input data structure of 1653 dimensions and received an agreement rate of 43\% on test dataset. In the year of 2017, they submitted a new article and designed a data structure with three dimensions of 5 by 34 by 5 just like an image to contain the tile information, 5 planes for the player's own hand tiles and four players' discarded tiles, and each plane has a data structure of 34 by 4. However, it just manages to contain a small part of the information available, leaving a large amount of information missing. This work achieved a test accuracy of 53.98\%~\cite{cnnmahjong}. 

\section{Data Structure Design}
\begin{figure}[tb]
  \centering
  \includegraphics[width = 3.5cm]{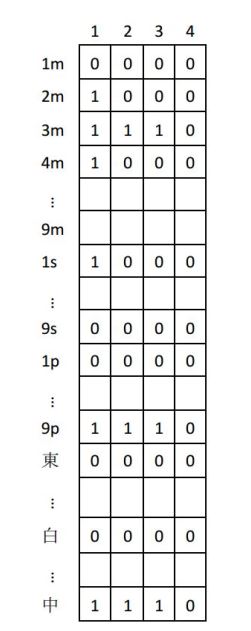}
  \caption{Data Plane Structure}
  \label{plane structure}
\end{figure}

In this paper, we propose a basic data structure which we call `plane'. It is a matrix with two dimensions, 34 in height and 4 in width. One plane structure is shown in~\figref{plane structure}. The game has 34 types of tiles and for each type, it will have in all four tiles on table, and this is how the height and width works. We believe data of this structure is suitable for network training.

Compared with previous one-hot structure in~\cite{cnnmahjong}, we believe it has at least three merits:
\begin{enumerate}
\item Firstly, it can have more potential and operating space. The only thing we can learn from the previous structure is the number of tiles in a given part, however in this structure, we can also be able to know some other hidden information, such as aka five information and the relations between the same kind of tiles. Although we do not choose to code aka information in this way, it still has potential in the future.
\item Secondly, as introduced later, we do not just simply contain the information about the tiles shown on table. We also design other planes to represent some other information, such as players' riichi information and wind tile information. For these kinds of irregular information, we need to code in a more different way, and our structure is obviously more suitable for coding.
\item Last but not least, it saves 20\% of the memory space. For deep learning tasks, more data usually leads to better training. One-fourth of the training data amount increase under limited hardware resources will be a huge improvement.
\end{enumerate}

For information on table, we divide them into different parts to represent them respectively, such as own hand tiles or players' discard tiles. For each divided part, we use one or several planes to represent respectively. To be in detail, for instance in~\figref{plane structure}, we have three 3m tiles in this specific part, so we have three ones filled in the 3m row, from the left to the right. Other rows are all filled in the same way. The left elements are all filled with zeros. In this approach, tile information can be represented properly.

In this section, we propose the data structure of in all 86 planes to represent the information available on table. The structure is shown in~\tabref{feature86}.
\begin{table}[tb]
\caption{INPUT FEATURES OF NEURAL NETWORKS}
\label{feature86}
\centering
\hbox to\hsize{\hfil
\begin{tabular}{l c} \hline\hline
    Feature & \# of planes\\ \hline
    Own hand tiles & 1\\
    Aka five mark & 1\\
    Discard tiles & 4\\
    Stealing tiles & 4\\
    Dora indicators & 1\\
    Riichi players & 3\\
    Rank & 4\\
    Kyoku & 8\\
    Round wind & 1\\
    Own wind & 1\\
    Past 1 situation & 13\\
    Past 2 situation & 9\\
    Past 3 situation & 9\\
    Past 4 situation & 9\\
    Past 5 situation & 9\\
    Past 6 situation & 9\\ \hline
\end{tabular}\hfil}
\end{table}

For information of present situation, we adopt first 28 planes to store, including two planes representing for one's own hand tiles together with aka five information, four planes for four players' discard tiles, four planes for four players' stealing tiles, one plane for dora indicators, three planes for other three players' riichi information, four planes for four players' rank information, eight planes for kyoku information of the game, one plane for the round wind and one plane for the own wind. The details are introduced below.

One player can have at most 14 closed tiles (which means not melded) in hand, therefore the first plane which represents for the player's own hand tiles will have at most 14 ones in it while it has in all 136 elements, and all other elements will be all filled with zeros. We add one more plane to mark the aka five information in hand. One aka five dora in hand will increase one han value once winning which is quite important, so usually it will not be easily discarded compared with normal five tiles. Players who have aka five doras in hand will usually try to build a legal hand pattern with these tiles, so we believe the aka five mark is necessary. One thing needs to be noticed is that although we fill the first plane with the basic plane style mentioned previously, for aka five information, we change the filling method a little bit. We let the whole corresponding row represent the aka five information, which means we four times enhances the number of tile information. For this second plane, if there appears one aka five tile in hand, we will fill the whole corresponding row of that tile type with four ones. In this way, we aim to maintain the feature that each type of tiles can appear at most four times on the table.

For stealing information, one player is able to call at most four melds in one game, with each meld containing three (pon or chi) or four (kan) tiles, so each stealing plane will have at most 16 ones in it, with all other elements filled with zeros. The filling method is same as explained in~\figref{plane structure}.

For dora indicator plane, each subgame will have one indicator showing to all at the beginning, and will appear at most four indicators during the game, since once the fourth kan is called which means the fifth dora indicator appears, it directly causes the end of the subgame according to game rules. If multiple dora indicators appear, they are usually in different types since the tile wall is formed randomly, but occasionally it may appear the same type. The filling method follows the basic coding rules same as before.

Besides, the riichi information of other players is also very important, since it is a clear announcement that some other players are already prepared to win and they have relatively large chances of big winnings, because they own a chance of hitting uradoras. The player should be especially careful not to discard dangerous tiles that are needed by riichi-declared players to avoid ron by them. These riichi planes contribute much on this task. Compared with models without riichi information planes contained in, the agreement rate of discard tile on test dataset raises over 1\%. The riichi plane here is coded in another way compared with previous planes. For previous ones, they are designed to store the number information of each type of tile for one part of information on table. Obviously, this is not a suitable coding method to represent riichi information. Therefore, here we adopt 0/1 planes for representation. A 0/1 plane is a plane that internal elements are filled with either all zeros or all ones, or we can recognize it as a pure white/black channel in CV tasks. This works well in our situation.

One player's playing style may change as the game continues. The player at top and the player at last may take different strategies, and the subgame at beginning and the subgame at last may also cause different reactions. So the score and kyoku information is an important factor for this game. However, the score information is hard to contain in our CNN model, even after preprocessing. We manage to take an approximation of score information as rank information. The coding method is the same as for riichi information. We have four planes to represent the player's rank, and the corresponding rank plane will be coded as a black channel. It is similar for kyoku information planes, and it will always have one black channel and seven white channels for the kyoku information.

For the round wind and the own wind information, in order to cause emphasis and maintain the principle of at most four tiles for each type of tile on the table, we mark the whole corresponding row with four ones to mark wind signs, which is the same coding approach of aka five information.

Past actions that players made also have great influence on present choice making. People can usually learn what other players need and not need from their past discard tiles. This is especially important for making a defense when another player declares riichi, because all players' passed discard tiles after that declaration are all safe, and tiles which are around his early discard tiles are usually safer than normal tiles. Although we already have planes of players' discard tiles already, it can only show what tiles players already discarded, but the discard orders cannot be learned. The closer the discard tiles are, normally more information they possess. After experiments, we include the last six rounds' information including own hand tiles, four players' discard tiles and four players' stealing tiles which means nine planes for each round. The number of dora indicators will increase by one during play if a kan is called by one player, hence last round's dora indicator information is also needed in order to mark a last round's kan sign. For riichi information, the round just after one's riichi is the most dangerous time for other players since it will add another one yaku and one han value if it wins during that round, so we also contain last round's riichi information to take care of that round. We include past information of dora indicators and other three players' riichi information only in past 1 situation but not others, this is why we have 13 planes for the past 1 situation but 9 planes each for others.

In all, we design a 86-plane data structure to represent information available on game table.

\section{Neural Network Structure Design}
In this section, we introduce our filter design for the CNN network. It has three convolutional blocks in the network, each one including one convolutional layer, one Batch Normalization layer and one Dropout layer.

Since our input is 34 by 4 in plane structure with 86 planes, our plane structure is too narrow with only 4 in width but 34 in height, therefore normal convolutional kernels such as 3x3 or 5x5 cannot be utilized well in our data structure. Finally we select three convolutional layers with 5 by 2 in kernel size and 100 in filter height. We adopt this filter three times with three convolutional layers, and we do not use any paddings in order to keep the matrix size no changed, which means the padding style is set to `VALID'.

A Batch Normalization layer and a Dropout layer of drop rate 0.5 are added after each convolutional layer for overfitting depression. We select ReLU as each block's activation function. We add one full-connected layer of 300 neurons after the flatten layer. We have a final output layer at last, designed after the full-connected layer but not the same for different tasks which are shown below.

\subsection{Discard Network}
In this paper, we evaluate the tile discard problem as a 34-class classification problem, since we have 34 types of tiles during the game, so the final layer becomes a softmax layer consisted of 34 neurons. A 14-class classification method is also another approach, because one player can have at most 14 tiles in hand thus the prediction output can just represent the order number of the tile in an ordered hand tile. However, after experiments we find that the 34-class one works much better. The only potential problem for the 34-class method is illegal discard tiles since we have 34 prediction choices for the network output but only at most 14 closed tiles in hand. However, as mentioned later in section~\uppercase\expandafter{\romannumeral8}, we find that all discard tile choices made by the network are legal predictions, showing surprisingly strong learning ability of our network.

\subsection{Pon Network}
One player can declare a pon when any other player discards one type of tile that he already has two closed ones in his own hand. The pon strategy decision can be recognized as a binary classification problem, while one represents for making the pon action and zero represents for not making that action. However, during real training for stealing network, unbalanced dataset becomes a problem. Among all pon-able situations, only about 30\% of the whole data makes a real pon, which means only 30\% of positive labels exist among the dataset. Usually for normal training problems, we can use the f1 score or other techniques to deal with unbalanced data in order to achieve good performances in all categories, however it becomes different for this task to some extent. The final goal for our task is to do a good job collaborating with the discard tile strategy for real play, but not just pursuing a relatively good result in all categories. A small ratio of positive labels means in real situations people tend not to do that decision. From this perspective of view, different loss functions and judgment levels act like representing for different styles while playing, for instance normal training will lead to an average style while f1-score-focused training may lead to a more offensive playing style. In this paper, we adopt the normal classification training judgment method at last.

\subsection{Chi Network}
Different from the pon decision, one player can only declare a chi with tiles discarded by the player left-side to form a legal sequence meld. It has a more severe ratio between positive and negative labels for chi situations, among which only about 14\% of the dataset has positive labels. Besides, not like pon, chi has three dealing methods with one tile, for that called tile can be either the smallest, largest or the middle part of the chi meld formed. Therefore, we design chi prediction as a 4-class classification problem, with zero representing action canceling, while one, two, and three standing for three types of chi actions with the called tile at the left, the middle and the right part of the meld formed.

\subsection{Kan Network}
For kan-able situations, we only have too small dataset since compared with pon and chi situations, kan becomes much rarer, and the ratio of declaring a kan is too low among the training dataset. Therefore, here we adopt a simplified strategy that we will not declare any open kans and will only declare a closed kan only when that kan is an isolated one which means that tile does not own any potential chances of involving in forming a chi meld according to the present hand tile pattern. So for kan decisions, we do not design a network for training, but make judgements by just a simple check.

\subsection{Riichi Network}
Besides stealing networks, we also build a network trained for riichi decision making. Of course the strategy of making instant riichi declaration once one player can is a more simple way, but it may cause problems because according to the game rules, once the player declares riichi, he cannot decide the discard tile by himself then, so it usually leads to dangerous situations. Besides, different hand patterns will lead to different scores, so sometimes waiting for several rounds to form a better winning pattern could also happen. The network structure is the same as the pon strategy network, since both of them are binary-classification problems.

\subsection{Winning Declaration}
Here we adopt a simplified strategy for winning declaration. Although the player has a chance to decide whether to announce winning when actually facing a winning situation, here for our program, we simplify this strategy by announcing the winning once the player can.

\section{Training Data and Experiment}
Our models and networks are trained on NVIDIA Tesla K10 GPU with \SI{32}{\giga\byte} memory size. We adopt the supervised machine learning method for prediction model and use game records collected from the `Houou' table at the biggest Japanese online mahjong site `Tenhou' in the year of 2015 as the training data. Only the players that are among top 0.1\% have the qualification of playing on the `Houou' table, so we believe these game records can have good enough quality for this record learning task. All game records from the `Houou' table are open for research and can be downloaded from the official website. During each subgame, we just follow one player for game record collection and make sampling until the player announces a riichi or winning declaration because once riichi declared, the player cannot determine one's discard tiles anymore, the discard tile is always the same as the drawn tile then. And in order for a better quality of the dataset, we will not learn the subgames that the player loses over 1500 scores.

\section{Training Results}
\subsection{Discard Network}
\begin{figure*}[tb]
\centering
\subfigure[Validation Accuracy]{
\begin{minipage}[t]{0.8\textwidth}
\centering
\includegraphics[width=12cm]{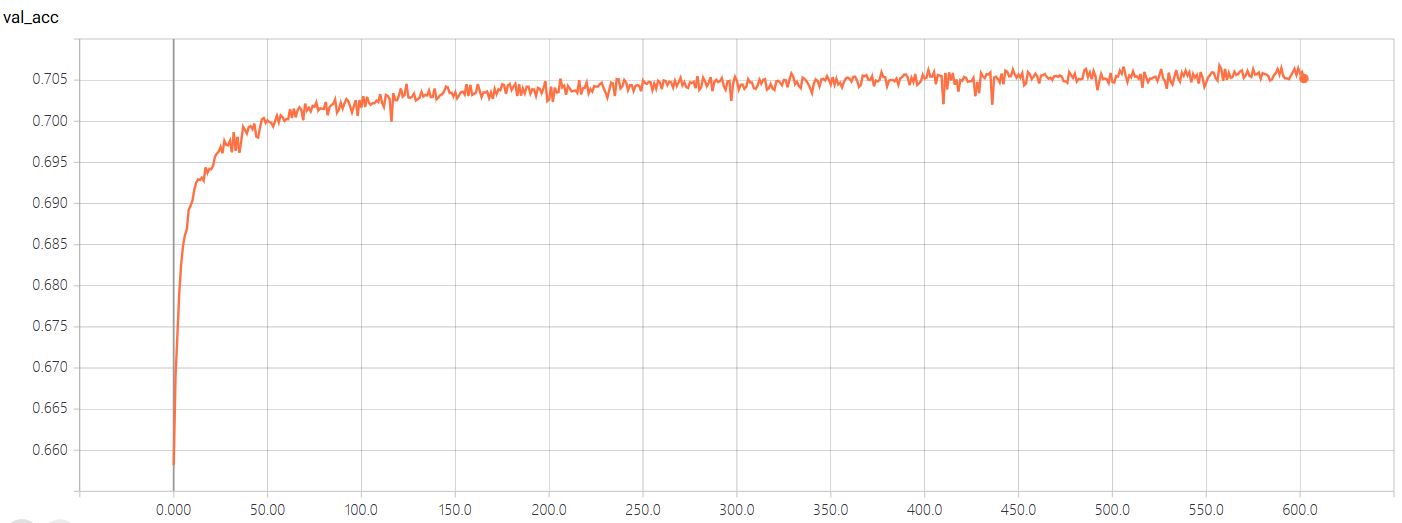}
\end{minipage}}
\subfigure[Validation Loss]{
\begin{minipage}[t]{\textwidth}
\centering
\includegraphics[width=12cm]{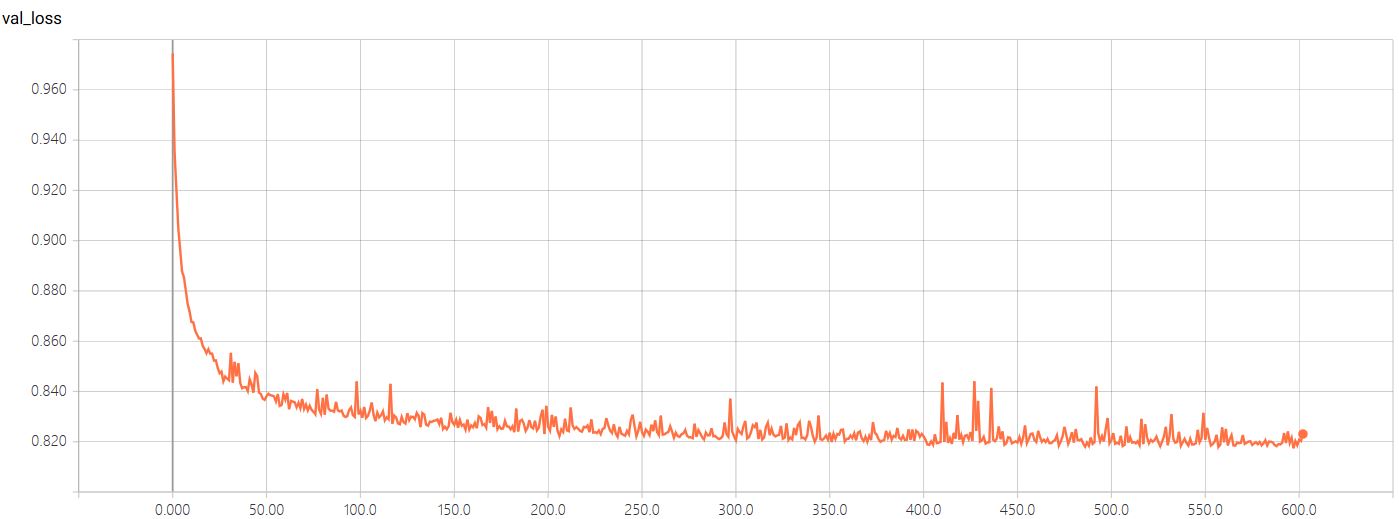}
\end{minipage}}
\caption{Training curve for tile discard strategy}
\label{Discard network curve}
\end{figure*}

We sample through all game records among the year of 2015 and obtain 2 135 331 different situations for training, with 10\% among it divided as the validation dataset. We set dropout rate at 0.5 and batch size at 256. The final validation accuracy reaches 70.71\%, which is shown in~\figref{Discard network curve}. For test data, we pick situations from the year of 2014 so the test dataset is totally different from training data. We picked 50 000 valid data, with at most one random situation from one subgame and without relevance between, and achieve an agreement rate accuracy of 70.44\%, while the state-of-art result is 62.1\%.

\subsection{Pon Network}
For the year of 2015, we obtain in all 212 845 valid situations for pon strategy training. Our validation accuracy arrives at 88.39\% during training. For the test dataset, we collect 30 000 data from the previous year of 2014. Test results are shown in \tabref{test pon}. The rows represent for real labels of True or False, while the columns represent for the predictions made by the neural network. We achieve a whole accuracy of 88.32\% on test data. We get precision of 81.46\% and recall of 78.21\%, leading to an f1 score of 0.798.
\begin{table}[tb]
\caption{TEST RESULTS ON PON NETWORK}
\label{test pon}
\centering
  \begin{tabular}{c|c c} \hline\hline
    & True & False \\ \hline
    True & 6923 & 1576 \\
    False & 1929 & 19572 \\ \hline
  \end{tabular}
\end{table}

\subsection{Chi Network}
For the year of 2015, we obtain in all 315 692 valid situations for chi strategy training. Our validation accuracy arrives at 90.79\% during training. For test, we collect 40 000 data from the previous year of 2014. Test results are shown in~\tabref{test chi}. The rows represent for real labels, while the columns represent for the prediction ones. As a 4-class classification problem, instead of True/False predictions, we make a prediction of 0, 1, 2 and 3. Here 0 means chi canceling, while the other three numbers indicate three types of chi mentioned before. We achieve a whole accuracy of 90.62\% on test data. We can also notice that, for real chi-called labels, to call which kind of chi reaches a very high accuracy of around 96.11\%, which means once the player decides to call a chi set, he can usually make the chi action type right. If we simplify it as a whether-to-chi-or-not binary classification problem, the accuracy will raise by another 0.3\%.
\begin{table}[tb]
\caption{TEST RESULTS ON CHI NETWORK}
\label{test chi}
\centering
\hbox to\hsize{\hfil
  \begin{tabular}{c|c c c c} \hline\hline
      & 0 & 1 & 2 & 3 \\ \hline
    0 & 33357 & 817 & 777 & 858 \\
    1 & 338 & 813 & 22 & 9 \\
    2 & 509 & 18 & 1289 & 29 \\
    3 & 335 & 7 & 32 & 790\\ \hline
  \end{tabular}\hfil}
\end{table}

\subsection{Riichi Network}
For the year of 2015, we obtain in all 99 999 valid situations for riichi strategy training. Our validation accuracy arrives at 77.18\% during training. For test, we collect 15 000 data from the previous year of 2014. Test results are shown in \tabref{test riichi}. The rows represent for real labels of True/False and the columns represent for the predictions the same as in pon strategy. We achieve a whole accuracy of 75.85\% on test data.
\begin{table}[tb]
\caption{TEST RESULTS ON RIICHI NETWORK}
\label{test riichi}
\centering
  \begin{tabular}{c|c c} \hline\hline
    & True & False \\ \hline
    True & 3017 & 1915 \\
    False & 1558 & 8510 \\ \hline
  \end{tabular}
\end{table}

\subsection{Evaluation on Tenhou}
We make our program playing matches on Tenhou sever against human players. The rule is set with East-South Battle, with Aka dora and open tanyao. Here we choose to play East-South Battles instead of East Battles because East Battles are usually too short so that fortune takes too much ratio. We double the length of the game in order to get a more precise performance judgment. After playing 300 matches, we reach a rating of around 1850, while an average intermediate player is around the rating of 1600, which means our program is much stronger than an intermediate player, even with the simple direct combination of several neural networks.

\section{Interesting Findings}
Pooling is recognized as a very useful and efficient tool for CNN network down-sampling, especially in computer vision tasks. However during experiment, we find that the pooling layer does not function well and its existence even decreases our result. Usually for computer vision tasks, image recognition has the property of movement invariance and rotation invariance, so pooling usually does a good job for down-sampling, having fewer parameters and forming more robust neural networks. However, for our task, the data plane structure is too elaborate and concise, and the down-sampling function of pooling will lose too much information and lead to a worse accuracy.

From the test cases, we also find that our proposed data structure makes a perfect understanding of the concept of hand tiles and what are the dora tiles only learning from the dora indicators. The discard tile prediction is a 34-class classification problem, however the player is just able to have at most 14 closed tiles in hand. Therefore, whether all predictions made out from the 34 classes are legal choices within the at most 14 types of tiles in hand at any situation is a meaningful question. Although designing another classification method is also an approach, for instance making predictions of which order is the tile that should be discarded in hand where the tiles are in sequential order. However, this does not make much sense and performs worth than our present model. For the result, within the 50 000 test cases, we find the predictions made by our neural network are all among the legal tile choices in hand, achieving a rate of 100\%. It is obvious that the network already learns the output range for tile discard prediction, which is an exciting finding. Besides, according to the game rules, dora tiles and dora indicators do not own a simple linear relationship but are divided into several loops as mentioned previously. However, our network understands that perfectly, and with the dora indicator plane contained in instead of directly using a plane to store dora information obtained from dora indicators, the agreement rate accuracy even raises about 1\%, which we think is because of the maintenance of the feature that at most four tiles of each type will appear on the table.

\section{Conclusion}
In this paper, we have proposed a new data model in three dimensions designed for the game of Mahjong, achieving a best performance so far on agreement rate of tile discarding after training on CNN networks. We simplified this prediction task as a multi-class classification problem and reached an agreement rate of 70.44\% while state-of-art accuracy is just 62.1\%. We also built an AI program playing online on `Tenhou' website for strength evaluation. We show that even by just simply combining several trained neural networks together without any human knowledge, our program reaches a rating of around 1850 after 300 matches, while the average rating for an intermediate player is just around 1600.

For future work, our model is still not perfect yet. We haven't included all information on table. Specific number information such as riichi stick number, honba number and score information is missing. We haven't found a suitable way to include this information in our model. We managed to contain rank information which is a rough approximation of score information, but there may have deviation between. Besides, since past actions made have great influences on present choice making, we included last 6 rounds' decision information in our model and obtained good results. However, a length period of six rounds may not always be enough. A perfect machine should be able to know all past actions. We have also tried sequential models of Gated Recurrent Unit (GRU) with invariant time-step training, although the result is better than the previous baseline but worse than our best result. Last but not least, for program strength evaluation, we made several simplifications for strategy decisions, such as kan and winning strategies, and just made a simple direct combination of several trained networks together. With more elaborately designed strategies for details and more specified knowledge of game rules and server rating calculation method, our program can absolutely reach a higher rating for strength evaluation.

\section*{Acknowledgment}
This work was supported by JST ERATO Grant Number JPMJER1501, Japan.

\bibliographystyle{IEEEtran}
\input{main.bbl}

\end{document}